\documentclass[runningheads]{llncs}
\usepackage{graphicx}
\usepackage{amsmath,amssymb} 
\usepackage{color}
\usepackage{placeins}

\usepackage{booktabs}
\usepackage{multirow}
\usepackage{subcaption}
\usepackage{float}

\DeclareMathOperator*{\argmax}{arg\,max}

\newcommand{\eg}{\emph{e.g.}}
\newcommand{\ie}{\emph{i.e.}}
\newcommand{\etal}{\emph{et al.}}

\usepackage[dvipsnames]{xcolor}
\definecolor{mygray}{gray}{0.}
\newcommand{\comp}[1]{\textcolor{mygray}{\emph{#1}}}

\begin{document}
\pagestyle{headings}
\mainmatter

\def\ACCV20SubNumber{656}  

\title{Self-Supervised Multi-View Synchronization Learning for 3D Pose Estimation} 
\titlerunning{Multi-View Synchronization Learning for 3D Pose Estimation}
%
\author{Simon Jenni\orcidID{0000-0002-9472-0425} \and 
Paolo Favaro\orcidID{0000-0003-3546-8247}}
\authorrunning{S. Jenni and P. Favaro}
%
\institute{University of Bern, Switzerland\\
\email{\{simon.jenni,paolo.favaro\}@inf.unibe.ch}}

\maketitle

\begin{abstract}
Current state-of-the-art methods cast monocular 3D human pose estimation as a learning problem by training neural networks on large data sets of images and corresponding skeleton poses. 
In contrast, we propose an approach that can exploit small annotated data sets by fine-tuning networks pre-trained via self-supervised learning on (large) unlabeled data sets. 
To drive such networks towards supporting 3D pose estimation during the pre-training step, we introduce a novel self-supervised feature learning task designed to focus on the 3D structure in an image. We exploit images extracted from videos captured with a multi-view camera system. The task is to classify whether two images depict two views of the same scene up to a rigid transformation. In a multi-view data set, where objects deform in a non-rigid manner, a rigid transformation occurs only between two views taken at the exact same time, \ie, when they are synchronized.
We demonstrate the effectiveness of the synchronization task on the Human3.6M data set and achieve state-of-the-art results in 3D human pose estimation. 
\end{abstract}

\section{Introduction}

\begin{figure}[t]
    \centering
    \begin{subfigure}{.49\linewidth}
    \includegraphics[width=\linewidth]{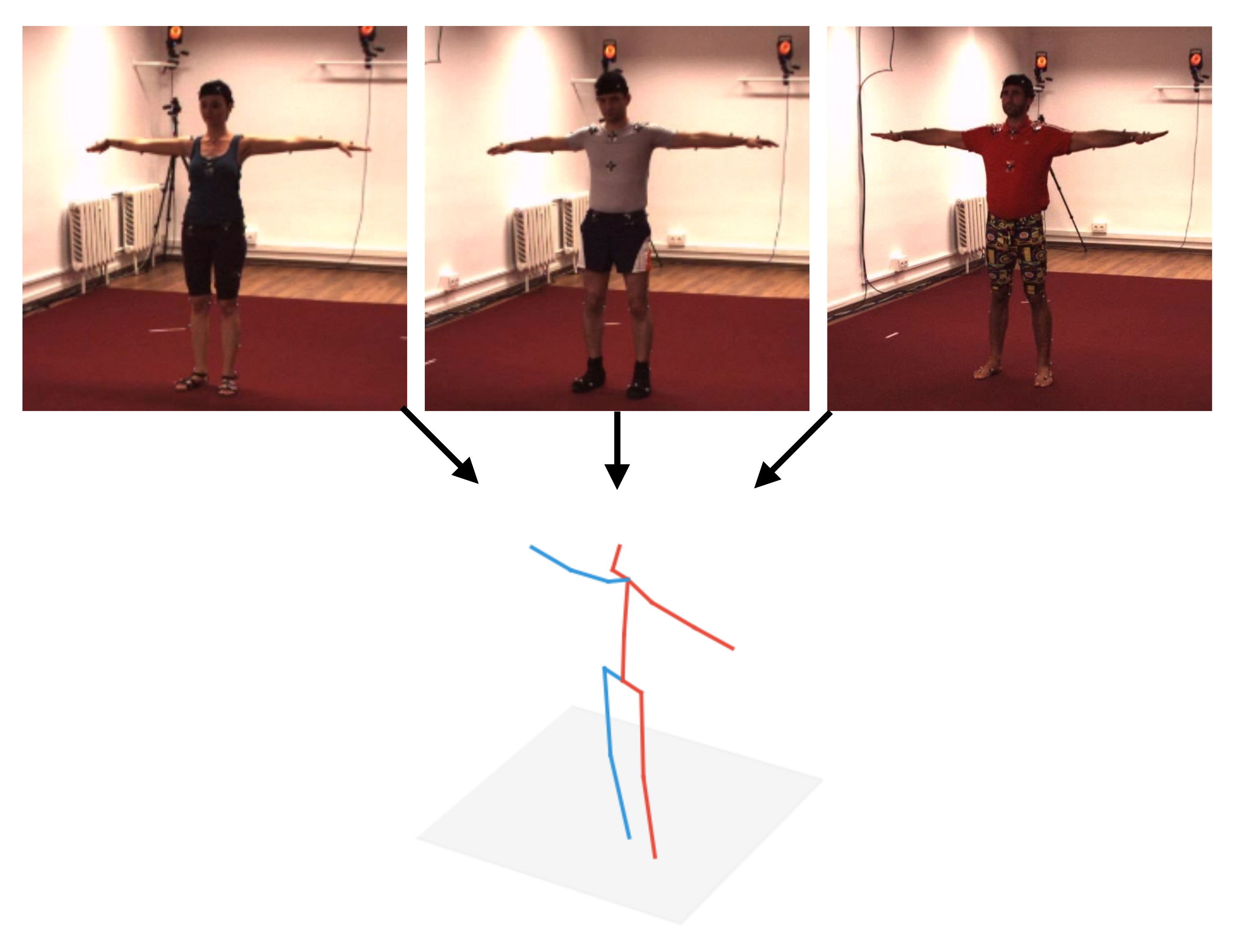}\\
    \subcaption[]{Desired subject invariance.}
    \end{subfigure}
    \centering
    \begin{subfigure}{.49\linewidth}
    \includegraphics[width=\linewidth]{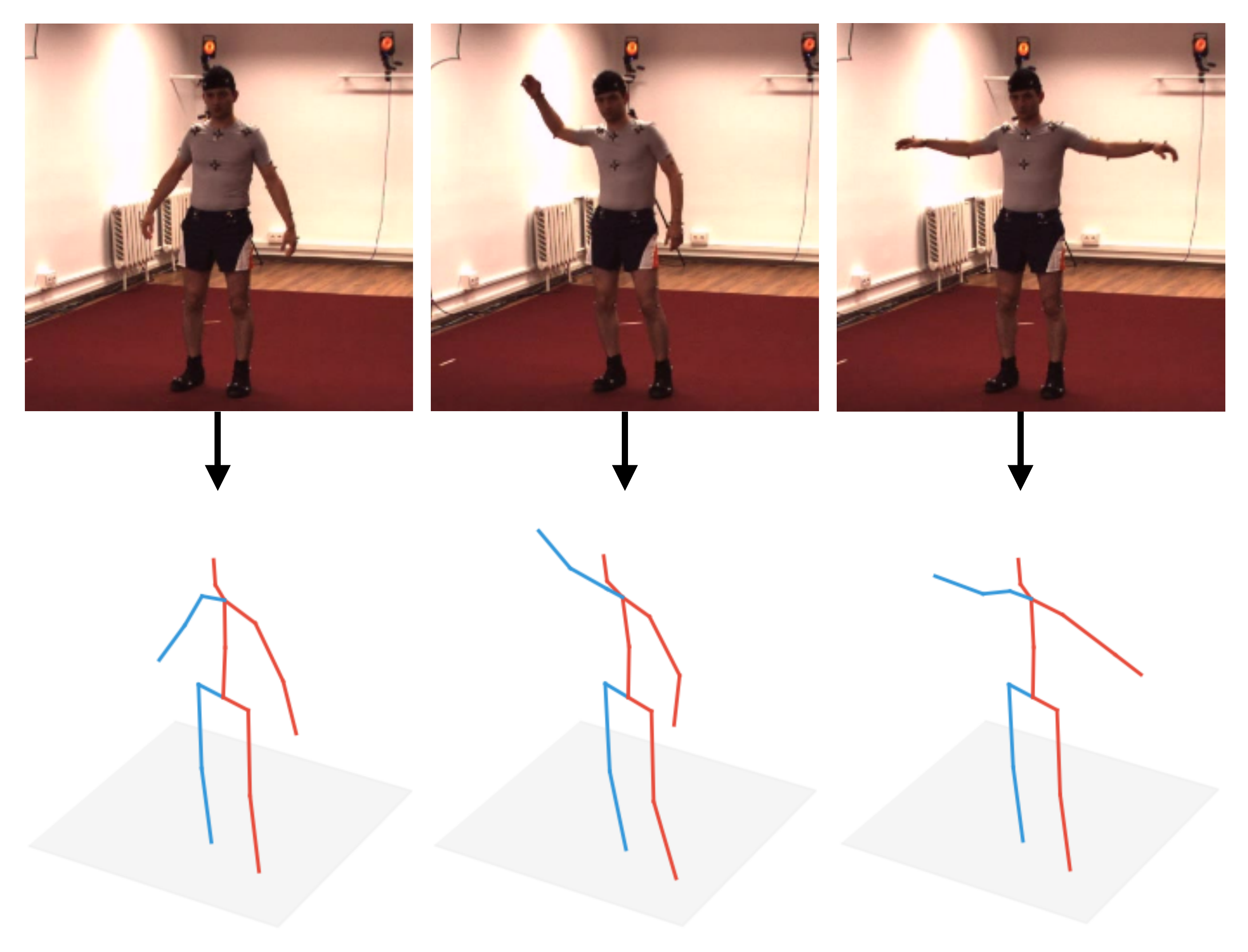}\\
    \subcaption[]{Desired pose equivariance.}
    \end{subfigure}
    \caption{\textbf{Monocular 3D human pose estimation.} An ideal regressor would be able to generalize across subjects regardless of their appearance, as shown in (a), and be sensitive to small pose variations, such as those shown in (b).}
    \label{fig:3DposeChallenges}
\end{figure}

The ability to accurately reconstruct the 3D human pose from a single real image opens a wide variety of applications including computer graphics animation, postural analysis and rehabilitation, human-computer interaction, and image understanding \cite{rosenhahn2008human}.
State-of-the-art methods for monocular 3D human pose estimation employ neural networks and require large training data sets, where each sample is a pair consisting of an image and its corresponding 3D pose annotation \cite{h36m_pami}. 
The collection of such data sets is expensive, time-consuming, and, because it requires controlled lab settings, the diversity of motions, viewpoints, subjects, appearance and illumination, is limited (see Fig.~\ref{fig:3DposeChallenges}).
Ideally, to maximize diversity, data should be collected in the wild. However, in this case precise 3D annotation is difficult to obtain and might require costly human intervention.

In this paper, we overcome the above limitations via \emph{self-supervised learning} (SSL). SSL is a method to build powerful latent representations by training a neural network to solve a so-called \emph{pretext} task in a supervised manner, but without manual annotation \cite{doersch2015unsupervised}. The pretext task is typically an artificial problem, where a model is asked to output the transformation that was applied to the data. 
One way to exploit models trained with SSL is to transfer them to some target task on a small labeled data set via \emph{fine-tuning} \cite{doersch2015unsupervised}. 
The performance of the transferred representations depends on how related the pretext task is to the target task. Thus, to build latent representations relevant to 3D human poses, we propose a pretext task that implicitly learns 3D structures.
To collect data suitable to this goal, examples from nature point towards multi-view imaging systems.
In fact, the visual system in many animal species hinges on the presence of two or multiple eyes to achieve a 3D perception of the world \cite{nityananda2017stereopsis,harland2012jumping}. 
3D perception is often exemplified by considering two views of the same scene captured at the same time and by studying the correspondence problem. Thus, we take inspiration from this setting and pose the task of determining if two images have been captured at exactly the same time. In general, the main difference between two views captured at the same time and when they are not, is that the former are always related by a rigid transformation and the latter is potentially not (\eg, in the presence of articulated or deformable motion, or multiple rigid motions). Therefore, we propose as pretext task the detection of \emph{synchronized views}, which translates into a classification of \emph{rigid} versus \emph{non-rigid} motion.

As shown in Fig.~\ref{fig:3DposeChallenges}, we aim to learn a latent representation that can generalize across subjects and that is sensitive to small pose variations. Thus, we train our model with pairs of images, where the subject identity is irrelevant to the pretext task and where the difference between the task categories are small pose variations. To do so, we use two views of the same subject as the synchronized (\ie, rigid motion) pair and two views taken at different (but not too distant) time instants as the unsynchronized (\ie, non-rigid motion) pair.\footnote{If the subject does not move between the two chosen time instants, the unsynchronized pair would be also undergoing a rigid motion and thus create ambiguity in the training. However, these cases can be easily spotted as they simply require detecting no motion over time. Besides standing still, the probability that a moving subject performs a rigid motion is extremely low. In practice, we found experimentally that a simple temporal sub-sampling was sufficient to avoid these scenarios.}
Since these pairs share the same subject (as well as the appearance) in all images, the model cannot use this as a cue to learn the correct classification. 
Moreover, we believe that the weight decay term that we add to the loss may also help to reduce the effect of parameters that are subject-dependent, since there are no other incentives from the loss to preserve them. 
Because the pair of unsynchronized images is close in time, the deformation is rather small and forces the model to learn to discriminate small pose variations.
Furthermore, to make the representation sensitive to left-right symmetries of the human pose, we also introduce in the pretext task as a second goal the classification of two synchronized views into \emph{horizontally flipped} or not.
This formulation of the SSL task allows to potentially train a neural network on data captured in the wild simply by using a synchronized multi-camera system. As we show in our experiments, the learned representation successfully embeds 3D human poses and it further improves if the background can also be removed from the images. 
We train and evaluate our SSL pre-training on the Human3.6M data set \cite{h36m_pami}, and find that it yields state-of-the-art results when compared to other methods under the same training conditions. We show quantitatively and qualitatively that our trained model can generalize across subjects and is sensitive to small pose variations.
Finally, we believe that this approach can also be easily incorporated in other methods to exploit additional available labeling (\eg, 2D poses). Code will be made available on our project page  \texttt{\url{https://sjenni.github.io/multiview-sync-ssl}}.

Our contributions are:
1) A novel self-supervised learning task for multi-view data to recognize when two views are synchronized and/or flipped; 
2) Extensive ablation experiments to demonstrate the importance of avoiding shortcuts via the static background removal and the effect of different feature fusion strategies;
3) 
State-of-the-art performance on 3D human pose estimation benchmarks.
    

\section{Prior work}

In this section, we briefly review literature in self-supervised learning, human pose estimation, representation learning and synchronization, that is relevant to our approach.\\
\noindent \textbf{Self-supervised learning.} Self-supervised learning is a type of unsupervised representation learning that has demonstrated impressive performance on image and video benchmarks. These methods exploit pretext tasks that require no human annotation, \ie, the labels can be generated automatically. Some methods are based on predicting part of the data \cite{pathak2016context,zhang2016colorful,noroozi2016unsupervised}, some are based on contrastive learning or clustering \cite{chen2020simple,he2020momentum,caron2018deep}, others are based on recognizing absolute or relative transformations \cite{jenni2020steering,gidaris2018unsupervised,zhang2019aet}. Our task is most closely related to the last category since we aim to recognize a transformation in time between two views.\\
\noindent \textbf{Unsupervised learning of 3D.} Recent progress in unsupervised learning has shown promising results for learning implicit and explicit generative 3D models from natural images  \cite{wu2020unsupervised,nguyen2019hologan,szabo2019unsupervised}. The focus in these methods is on modelling 3D and not performance on downstream tasks. Our goal is to learn general purpose 3D features that perform well on downstream tasks such as 3D pose estimation. \\
\noindent \textbf{Synchronization.} Learning the alignment of multiple frames taken from different views is an important component of many vision systems. Classical approaches are based on fitting local descriptors \cite{agarwala2005panoramic,sand2004video,tuytelaars2004synchronizing}. More recently, methods based on metric learning \cite{wieschollek2017learning} or that exploit audio \cite{liang2017synchronization} have been proposed. We provide a simple learning based approach by posing the synchronization problem as a binary classification task. Our aim is not to achieve synchronization for its own sake, but to learn a useful image representation as a byproduct.\\
\noindent \textbf{Monocular 3D pose estimation.} State-of-the-art 3D pose estimation methods make use of large annotated in-the-wild 2D pose data sets \cite{andriluka20142d} and data sets with ground truth 3D pose obtained in indoor lab environments. We identify two main categories of methods: 1) Methods that learn the mapping to 3D pose directly from images \cite{li20143d,li2015maximum,tekin2016structured,zhou2016deep,tekin2017learning,pavlakos2017harvesting,pavlakos2017coarse,mehta2017monocular,sun2018integral} often trained jointly with 2D poses \cite{popa2017deep,tome2017lifting,mehta2017vnect,rogez2017lcr,dabral2018learning}, and 2) Methods that learn the mapping of images to 3D poses from predicted or ground truth 2D poses \cite{martinez2017simple,zhou2017towards,moreno20173d,rayat2018exploiting,fang2018learning,chen20173d,zhao2019semantic,sharma2019monocular,wang20193d}. To deal with the limited amount of 3D annotations, some methods explored the use of synthetic training data \cite{chen2016synthesizing,rogez2016mocap,varol2017learning}. In our transfer learning experiments, we follow the first category and predict 3D poses directly from images. However, we do not use any 2D annotations. \\
\noindent \textbf{Weakly supervised methods.} Much prior work has focused on reducing the need for 3D annotations. One approach is weak supervision, where only 2D annotation is used. These methods are typically based on minimizing the re-projection error of a predicted 3D pose \cite{kocabas2019self,rhodin2018learning,chen2019weakly,chen2019unsupervised,kanazawa2018end,wandt2019repnet,pavllo20193d}. To resolve ambiguities and constrain the 3D, some methods require multi-view data \cite{kocabas2019self,rhodin2018learning,chen2019weakly}, while others rely on unpaired 3D data used via adversarial training \cite{kanazawa2018end,wandt2019repnet}. \cite{chen2019unsupervised} solely relies on 2D annotation and uses an adversarial loss on random projections of the 3D pose. 
Our aim is not to rely on a weaker form of supervision (\ie, 2D annotations) and instead leverage multi-view data to learn a representation that can be transferred on a few annotated examples to the 3D estimation tasks with a good performance. \\
\noindent \textbf{Self-supervised methods for 3D human pose estimation.} Here we consider methods that do not make use of any additional supervision, \eg, in the form of 2D pose annotation. These are methods that learn representations on unlabelled data and can be transferred via fine-tuning using a limited amount of 3D annotation. 
Rhodin \etal~\cite{rhodin2018unsupervised} learn a 3D representation via novel view synthesis, \ie, by reconstructing one view from another. Their method relies on synchronized multi-view data, knowledge of camera extrinsics and background images. Mitra \etal~\cite{mitra2020multiview} use metric learning on multi-view data. The distance in feature space for images with the same pose, but different viewpoint is minimized while the distance to hard negatives sampled from the mini-batch is maximized. By construction, the resulting representation is view invariant (the local camera coordinate system is lost) and the transfer can only be performed in a canonical, rotation-invariant coordinate system. We also exploit multi-view data in our pre-training task. In contrast to \cite{rhodin2018unsupervised}, our task does not rely on knowledge of camera parameters and unlike \cite{mitra2020multiview} we successfully transfer our features to pose prediction in the local camera system. 

\section{Unsupervised learning of 3D pose-discriminative features}

Our goal is to build image features that allow to discriminate different 3D human poses.
We achieve this objective by training a network to detect if two views depict the exact same scene up to a rigid transformation. 
We consider three different cases: 1) The two views depict the same scene at exactly the same time (rigid transformation); 2) The two views show the same scene, but at a different time (non-rigid transformation is highly likely); 3) The two views show the same scene, but one view is horizontally mirrored (a special case of a non-rigid transformation), which, for simplicity, we refer to as \emph{flipped}. 

To train such a network, we assume we have access to synchronized multi-view video data. The data set consists of $N$ examples (in the Human3.6M data set $N=S\times A$, where $S$ is the number of subjects and $A$ the number of actions) $\{x^{(i)}_{\nu,t}\}_{i=1,\ldots,N}$, where $\nu \in{\cal V}^{(i)}=\{1,\ldots,\nu^{(i)}\}$ indicates the different views of the $i$-th example and $t\in{\cal T}^{(i)} = \{1,\ldots,t^{(i)}\}$ indicates the time of the frame. Let $F$ denote the neural network 
trained to solve the self-supervised task. We will now describe the different self-supervision signals.


\subsection{Classifying synchronized and flipped views}

To train our neural network we define three types image pairs, each corresponding to a different 3D deformation: 
\begin{description}
    \item[Synchronized pairs.] In this case, the scenes in the two views are related by a rigid transformation. This is the case when the two images are captured at the same time instant, \ie, they are synchronized. 
    The input for this category is a sample pair $\mathbf{x}_p = (x^{(i)}_{\nu_1,t}, x^{(i)}_{\nu_2,t})$, where $i\in\{1,\ldots,N\}$, $\nu_1 \neq \nu_2 \in {\cal V}^{(i)}$, and $t\in {\cal T}^{(i)}$, \ie, a pair with different views taken at the same time.\\

    \item[Unsynchronized pairs.] 
    Pairs of images that are captured at different times (\ie, unsynchronized) are likely to undergo a non-rigid deformation (by assuming that objects are non static between these two time instants). To create such image pairs we sample  $\mathbf{x}_n = (x^{(i)}_{\nu_1,t_1}, x^{(i)}_{\nu_2,t_2})$, where $i\in\{1,\ldots,N\}$, $\nu_1 \neq \nu_2\in {\cal V}^{(i)}$, and $t_1, t_2\in {\cal T}^{(i)}$ such that $d_{min} <  |t_2 - t_1| < d_{max}$, where $d_{min}$ and $d_{max}$ define the range in time for sampling unsynchronized pairs. In our experiments, we set $d_{min}=4$ and $d_{max}=128$ and sample uniformly within this range.\\
    
    \item[Flipped pairs.] The last class of image pairs consists of images from two views which are synchronized, but where one of them has been horizontally mirrored. Let $\bar{x}$ denote the image obtained by applying horizontal mirroring to the sample $x$. Flipped pairs are defined as
    $\mathbf{x}_f = (x^{(i)}_{\nu_1,t}, \bar{x}^{(i)}_{\nu_2,t})$ or $\mathbf{x}_f = (\bar{x}^{(i)}_{\nu_1,t}, x^{(i)}_{\nu_2,t})$, where $i\in \{1,\ldots,N\}$, $\nu_1 \neq \nu_2\in {\cal V}^{(i)}$ and $t\in {\cal T}^{(i)}$.
    In the case of approximately symmetric objects (such as with human bodies) and in the absence of background cues (since we mirror the entire image) distinguishing this relative transformation between views requires accurate 3D pose discriminative features.
    
\end{description}

Although both \textbf{unsynchronized} and \textbf{flipped} pairs exhibit non-rigid deformations, they have distinct characteristics. In the case of flipped pairs, the 3D pose in the second view is heavily constrained by the one in the first view. In contrast, given a non negligible temporal gap between the frames, the 3D pose is much less constrained in the case of unsynchronized views.
\begin{figure}[t]
    \centering
    \includegraphics[width=\linewidth]{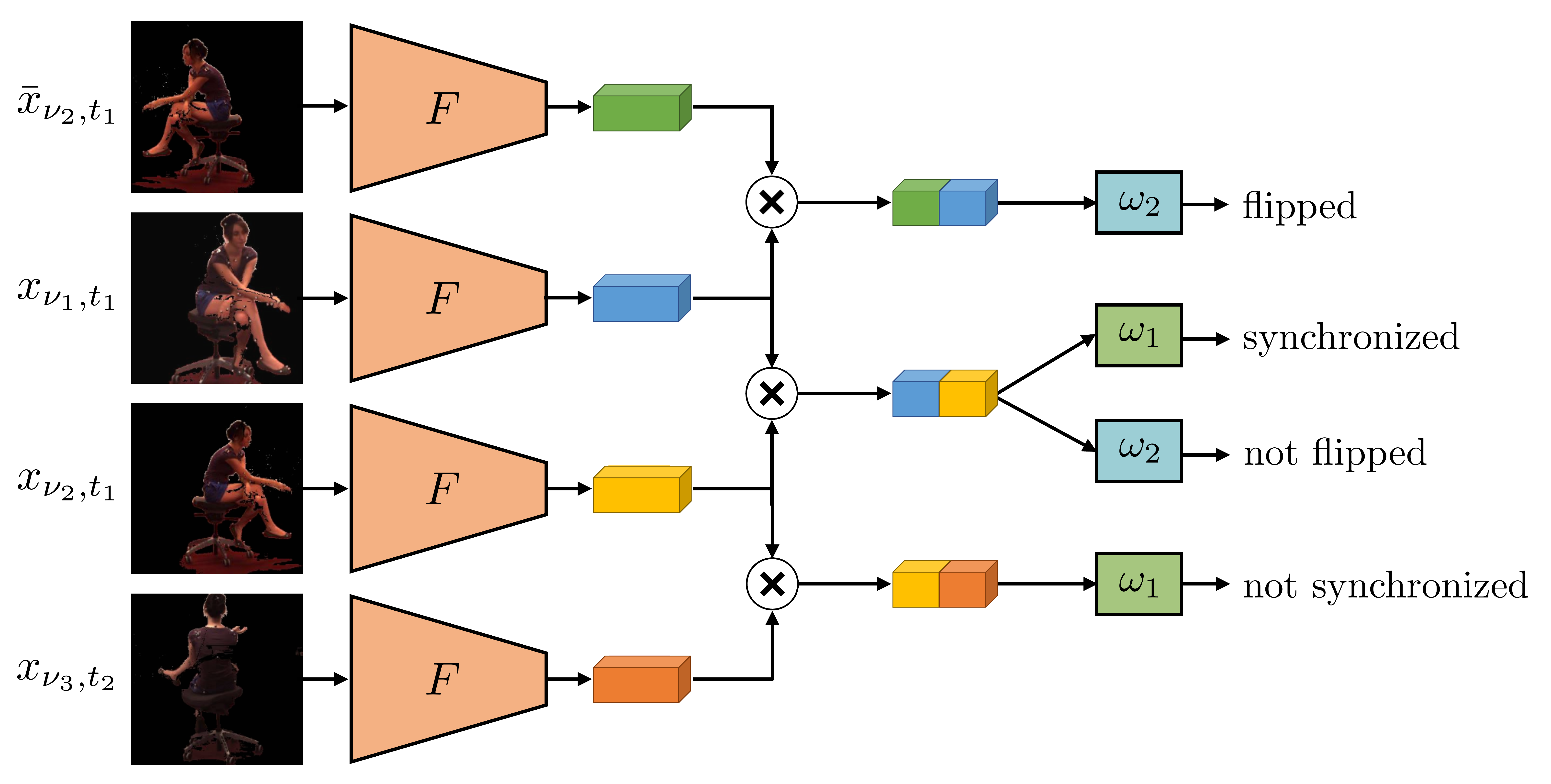}
    \caption{\textbf{An overview of the proposed self-supervised task.} We train a network $F$ to learn image representations that transfer well to 3D pose estimation by learning to recognize if two views of the same scene are \emph{synchronized} and/or \emph{flipped}. Our model uses a Siamese architecture. The inputs are  pairs of frames  of the same scene under different views. Frames are denoted by $x_{\nu,t}$, where $\nu$ indicates the viewpoint and $t$ the time ($\Bar{x}_{\nu,t}$ denotes a flipped frame). Pairs are classified into whether the frames are \emph{synchronized} or \emph{not synchronized}, and whether one of the frames was \emph{flipped} or not. }
    \label{fig:model}
\end{figure}

We define our self-supervised learning task as a combination of the classification of image pairs into \emph{synchronized vs. unsynchronized} and into \emph{flipped vs. not flipped}. Let $F_1(\mathbf{x})$ denote the predicted probability of synchronization in $\mathbf{x}$ and $F_2(\mathbf{x})$ the predicted probability of no flipping in $\mathbf{x}$ (both implemented via a sigmoid activation). The final loss function is then given by 
\begin{equation}
    \mathcal{L}_{SSL}=-\sum_{j} \log F_1\big(\mathbf{x}_p^{(j)}\big)F_2\big(\mathbf{x}_p^{(j)}\big) + \log\left(1-F_1\big(\mathbf{x}_n^{(j)}\big)\right) + \log\left(1-F_2\big(\mathbf{x}_f^{(j)}\big)\right).
\end{equation}
An illustration of the 
training is shown in Fig.~\ref{fig:model}.
Note that although $\mathbf{x}_f$ is a synchronized pair, we do not optimize the synchronization head $F_1$ on these examples. The reason for leaving this term undefined is that we equate synchronization to rigid motion, which does not hold in the case of $\mathbf{x}_f$. 

\subsection{Static backgrounds introduce shortcuts}

A common issue with self-supervised learning is that neural networks exploit low-level cues to solve the pretext task, when allowed to do so. In this case, they often do not learn semantically meaningful features. Such shortcuts have been observed in prior work, where, for instance, chromatic aberration \cite{pathak2016context} was used as a cue to solve a localization task. 

In our self-supervised learning task we observed that the shared image background in the data set can be used as a shortcut. This is most evident in the flipping task: Because the background is shared among all examples in the data set, it is easier to use texture in the background to determine image flipping. As a result, the network focuses on the background (instead of the foreground object of interest) with a decrease in its generalization performance. 
It is further possible that the network learns to detect and associate the absolute position of the person by focusing on the background as a cue for the synchronization task. 

We propose two solutions: One is \emph{background removal} and the other is \emph{background substitution}. Since the data set consists of non-moving cameras, we compute the background as the median image per view. Background removal is then performed via simple per-pixel thresholding. Although the resulting separation is not of high quality, we found it sufficient for our purpose.
Similarly, background substitution introduces an additional step where the background from a random view is used to replace the original (removed) one. Both approaches are evaluated in our experiments.
We expect that background removal or substitution would not be necessary for data captured in the wild with synchronized moving cameras or with a large variety of backgrounds.

\subsection{Implementation}

We implement the network $F$ using a Siamese architecture. As a shared backbone architecture $\Phi$ we use standard ResNet architectures \cite{he2016deep}. Samples $\textstyle(x^{(j)}_{\nu_1,t_1}, x^{(j)}_{\nu_2,t_2})$ are hence encoded into features vectors $(\Phi(x^{(j)}_{\nu_1,t_1}), \Phi(x^{(j)}_{\nu_2,t_2}))$. In our experiments we use the output of the global average pooling as the feature representation, \ie, $\Phi(x) \in \mathbb{R}
^{2048}$. To fuse the features we choose to use element-wise multiplication followed by a ReLU activation. We found this fusion mechanism to perform better than feature concatenation, which is often used in prior work \cite{pathak2016context,noroozi2016unsupervised,zhang2019aet}. The fused feature is fed as input to linear layers $\omega_1$ and $\omega_2$ to produce the final outputs. To summarize, we define $F_i(\mathbf{x}
)= \omega_i( \text{ReLU} ( \Phi(x^{(j)}_{\nu_1,t_1})\odot \Phi(x^{(j)}_{\nu_2,t_2})) )$, where $\odot$ denotes the  element-wise product.

The training images use the standard image resolution for ResNet architectures of  $224\times224$ pixels. To compare fairly with prior work, we extract crops centered on the subject. As data-augmentation, we add random horizontal flipping applied consistently to both frames in the input pair (this operation is performed before preparing the flipped pairs).

\subsection{Transfer to 3D human pose estimation}

We transfer the learned image representation to 3D human pose regression via fine-tuning. The training set in this case consists of a set of image\footnote{We drop the subscripts $\nu, t$ indicating viewpoint and time in this notation.} and target pairs $\{(x^{(i)}, y^{(i)})\}_{i=1,\ldots,N_p}$ with target 3D joint positions $y^{(i)}\in \mathbb{R}^{n_j\times3}$  represented in the local camera coordinate system. The number of joints is set to $n_j=17$ in our experiments on Human3.6M. 
As in prior work \cite{rhodin2018learning,rhodin2018unsupervised,mitra2020multiview}, we fix the root joint (at the pelvis) to zero. We effectively only regress the 16 remaining joints and correct for this in evaluation metrics by scaling the per joint errors by $\frac{17}{16}$.
We normalize the target poses for training using the per-joint mean and standard deviation computed on the training set as proposed in \cite{sun2018integral}. During evaluation, we un-normalize the predicted pose accordingly.
To predict the 3D pose we simply add a single linear layer $\omega_p$ on the feature encoding $\Phi(x^{(i)})$ resulting in our normalized prediction $P(x^{(i)})=\omega_p(\Phi(x^{(i)}))$. 
The network $P$ is trained to minimize the mean-squared error on the training set, \ie, $\mathcal{L}_{pose}=\frac{1}{N_p} \sum_{i=1}^{N_p} \|P(x^{(i)})-y^{(i)}\|_2^2$.

We again extract crops centered on the subject as in prior work \cite{rhodin2018learning,rhodin2018unsupervised,mitra2020multiview}. Since this corresponds to a change of the camera position, we correct for this by rotating the target pose accordingly, \ie, to virtually center the camera on the root joint.  As a form of data augmentation, we apply random horizontal flipping jointly to the input images and corresponding target 3D poses.


\section{Experiments}

\noindent \textbf{Dataset.} We perform an extensive experimental evaluation on the Human3.6M data set \cite{h36m_pami}. The data set consists of 3.6 million 3D human poses with the corresponding images. The data is captured in an indoor setting with 4 synchronized and fixed cameras placed at each corner of the room. Seven subjects perform 17 different actions. As in prior work, we use the five subjects with the naming convention S1, S5, S6, S7, and S8 for training and test on the subjects S9 and S11. We filter the training data set by skipping frames without significant movement. On average, we skip every fourth frame in the training data set. On the test set, we follow prior work and only evaluate our network on one every 64 frames.  

\noindent \textbf{Metrics.} To evaluate our method on 3D human pose regression we adopt the established metrics. We use the Mean Per Joint Prediction Error (MPJPE) and its variants Normalized MPJPE (NMPJPE) and Procrustes MPJPE (PMPJPE). In NMPJPE the predicted joints are aligned to the ground truth in a least squares sense with respect to the scale. PMPJE uses Procrustes alignment to align the poses both in terms of scale and rotation. 

\subsection{Ablations}
We perform ablation experiments to validate several design choices for our self-supervised learning task. We illustrate the effect of background removal with respect to the shortcuts in the case of non-varying backgrounds (as is the case for Human3.6M). We also demonstrate the effects of the two self-supervision signals (flipping and synchronization) by themselves and in combination and explore different feature fusion strategies. 

The baseline model uses background removal, the combination of both self-supervision signals and element-wise multiplication for feature fusion. 
The networks are pre-trained for 200K iterations on all training subjects (S1, S5, S6, S7, and S8) using our SSL task and without using annotations. We use a ResNet-18 architecture \cite{he2016deep} and initialize with random weights (\ie, we do not rely on ImageNet pre-training). 
Transfer learning is performed for an additional 200K iterations using only subject S1 for supervision. We freeze the first 3 residual blocks and only finetune the remaining layers.
Training of the networks was performed using the AdamW optimizer \cite{loshchilov2018decoupled} with default parameters and a weight decay of $10^{-4}$. We decayed the learning rate from $10^{-4}$ to $10^{-7}$ over the course of training using cosine annealing \cite{loshchilov2016sgdr}. The batch size is set to 16.

\begin{table}[t]
\centering
\caption{\textbf{Ablation experiments.} We investigate the influence of scene background  (a)-(d), the influence of the different self-supervision signals (e)-(g), the use of different fusion strategies (h)-(k), and the importance of synchronized multiview data (l) and (m). The experiments were performed on Human3.6M using only subject S1 for transfer learning.\\ \label{tab:ablations}}

\begin{tabular}{@{}c@{}l@{}c@{}c@{}c@{}}
\toprule
\phantom{o}\phantom{o}\phantom{o}\phantom{o} & \textbf{Ablation}\phantom{o} & \phantom{o}MPJPE\phantom{o} & \phantom{o}NMPJPE\phantom{o} & \phantom{o}PMPJPE \\ \midrule
(a) & Random with background & 167.8 & 147.1 & 124.5 \\  
(b) & Random without background & 165.5 & 145.4 & 114.5 \\  
(c) & SSL with background &  138.4  &  128.1   &  100.8  \\ 
(d) & SSL without background &  104.9 & 91.7 & 78.4 \\    
\midrule
(e) & Only flip & 129.9 & 117.6 & 101.1 \\ 
(f) & Only sync & 126.7 & 115.2 & 91.8 \\  
(g) & Both sync \& flip & 104.9 & 91.7 & 78.4 \\ 
\midrule 
(h) & Fusion concat  & 180.0  &  162.5  &  130.5  \\  
(i) & Fusion add  &  169.4  &  158.4  &  122.8  \\  
(j) & Fusion diff  &  108.2  &  93.9  &  80.4  \\  
(k) & Fusion mult  & 104.9 & 91.7 & 78.4  \\ 
\midrule
(l) &  Single-view SSL  &  158.3  &  142.0  &  106.9 \\  
(m) &  Multi-view SSL  & 104.9 & 91.7 & 78.4   \\
\bottomrule
\end{tabular}
\end{table}

We perform the following set of ablation experiments and report transfer learning performance on the test set in Table \ref{tab:ablations}:

\begin{description}
	\item [(a)-(d) Influence of background removal:] We explore the influence of background removal on the performance of randomly initialized networks and networks initialized with weights from our self-supervised pre-training. We observe that background removal provides only a relatively small gain in the case of training from scratch. The improvement in the case of our self-supervised learning pre-training is much more substantial. This suggests that  the static background introduces shortcuts for the pretext task;
	\item [(e)-(f) Combination of self-supervision signals:] 
	We compare networks  initialized by pre-training (e) only to detect flipping, (f) only to detect synchronization, and (g) on the combination of both. We observe that the synchronization on its own leads to better features compared to flipping alone. This correlates with the pretext task performance, where we observe that the accuracy on the flipping task is higher than for synchronization. Interestingly, the combination of both signals gives a substantial boost. This suggests that both signals learn complementary features;
	\item [(h)-(k) Fusion of features:] Different designs for the fusion of features in the Siamese architecture are compared. We consider (h) concatenation, (i) addition, (j) subtraction, and (k) element-wise multiplication. We observe that multiplication performs the best. 
	Multiplication allows a simple encoding of similarity or dissimilarity in feature space solely via the sign of each entry. The fused feature is zero (due to the ReLU) if the two input features do not agree on the sign;
	\item [(l)-(m) Necessity of multi-view data:] We want to test how important synchronized multi-view data is for our approach. We compare to a variation of our task, where we do not make use of multi-view data. Instead, we create artificial synchronized views via data augmentation. The images in the input pairs are therefore from the same camera view, but with different augmentations (\ie, random cropping, scaling and rotations). This corresponds to using a 2D equivalent of our self-supervised learning task. We observe drastically decreased feature performance in this case.
\end{description}


\subsection{Comparison to prior work}

\begin{table}[t]
\centering
\caption{\textbf{Comparison to prior work.} We compare to other prior work. 
All methods are pre-trained on all the training subjects of Human3.6M. Transfer learning is performed either using all the training subjects or only S1 for supervision. Methods using large amounts of data with 2D pose annotation are \comp{italicized}. * indicates the use of an ImageNet pretrained ResNet-50.  $^\dagger$ indicates a viewpoint invariant representation of the 3D pose.\\
\label{tab:comparison}}
\begin{tabular}{@{}c@{}l@{}c@{}c@{}c@{}}
\toprule
\textbf{Supervision}\phantom{o}\phantom{o}             & \textbf{Method}\phantom{o}            & \phantom{o}MPJPE\phantom{o} & \phantom{o}NMPJPE\phantom{o} & \phantom{o}PMPJPE \\ 
\midrule
\multirow{10}{*}{\textbf{All}}   & \comp{Chen et al.\cite{chen2019weakly}*}  & \comp{80.2} & \comp{-} & \comp{58.2} \\
& \comp{Kocabas et al.\cite{kocabas2019self}*}  & \comp{51.8} & \comp{51.6} & \comp{45.0} \\

& \comp{Rhodin et al.\cite{rhodin2018learning}*}  & \comp{66.8} & \comp{63.3} & \comp{51.6} \\
                        & Rhodin \etal \cite{rhodin2018learning}*  & - & 95.4 & - \\
                        & Rhodin (UNet) \etal \cite{rhodin2018unsupervised}  & - & 127.0 & - \\
                        & Rhodin \etal \cite{rhodin2018unsupervised}*  & - & 115.0 & - \\
                        & Mitra \etal \cite{mitra2020multiview}*$^\dagger$       & 94.3 & 92.6 & 72.5 \\
                        & Ours              & 79.5 & 73.4 & 59.7 \\  
                        & Ours*              & 72.6 & 68.5 & 54.5  \\ 
                        & Ours* (with background)  & 64.9 & 62.3 & 53.5 \\  
\midrule

\multirow{10}{*}{\textbf{S1}}    & \comp{Chen et al.\cite{chen2019weakly}*}  & \comp{91.9} & \comp{-} & \comp{68.0} \\
& \comp{Kocabas et al.\cite{kocabas2019self}*}  & \comp{-} & \comp{67.0} & \comp{60.2} \\
 & \comp{Rhodin et al.\cite{rhodin2018learning}*}  & \comp{-} & \comp{78.2} & \comp{-} \\
 & Rhodin \etal \cite{rhodin2018learning}*  & - & 153.3 & 128.6 \\
                        & Rhodin (UNet) \etal \cite{rhodin2018unsupervised}  & 149.5 & 135.9 & 106.4 \\
                        & Rhodin \etal \cite{rhodin2018unsupervised}*  & 131.7 & 122.6 & 98.2 \\
                        & Mitra \etal \cite{mitra2020multiview}*$^\dagger$       & 121.0 & 111.9 & 90.8 \\
                        & Ours  &  104.9 & 91.7 & 78.4  \\ 
                        & Ours*  & 101.2 & 89.6 & 76.9 \\ 
                        & Ours* (with background)  & 101.4 & 93.7 & 82.4 \\  

\bottomrule
\end{tabular}
\end{table}

We compare our method to prior work by Rhodin \etal~\cite{rhodin2018learning,rhodin2018unsupervised} and Mitra \etal~\cite{mitra2020multiview} on Human3.6M.
To the best of our knowledge, these are the only methods that use the same training settings as in our approach. Other methods, which we also report, train their networks with additional weak supervision in the form of 2D pose annotation (see Table~\ref{tab:comparison}). 

We pretrain the networks on our self-supervised learning task for 200K iterations on all training subjects (namely, S1, S5, S6, S7, and S8). Our comparison uses two different protocols: 1) \textbf{All}, where transfer learning is performed on all the training subjects and 2) \textbf{S1}, where only subject S1 is used for transfer learning.   
We use two different network architectures: 1) A ResNet-18 initialized with random weights, \ie, no ImageNet pre-training, and 2) A ResNet-50 initialized with ImageNet pre-trained weights.
Since prior work transfers to images with backgrounds we include results with a ResNet-50 also fine-tuned on images with backgrounds. To reduce the domain gap between pre-training and transfer and to eliminate shortcuts, we pre-train on both images, where backgrounds are removed, and images with substituted backgrounds. We keep the ratio without and with substituted backgrounds to 50:50 during pre-training.  
During transfer learning we freeze the first 3 residual blocks and fine-tune the remaining layers for 200K iterations using a mini-batch size of 16. 
Training of the networks was again performed using the AdamW optimizer \cite{loshchilov2018decoupled} with default parameters, initial learning rate of $10^{-4}$ with cosine annealing, and a weight decay of $10^{-4}$. 

\begin{figure}[t]
    \centering
    \includegraphics[height=0.15\linewidth,trim={0px 0px 0px 0px},clip]{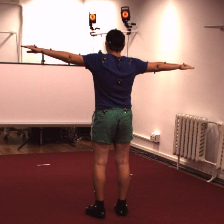}
    \includegraphics[height=0.15\linewidth,trim={0px 0px 0px 0px},clip]{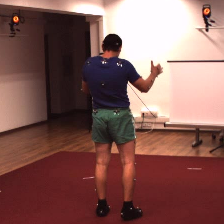}
    \includegraphics[height=0.15\linewidth,trim={0px 0px 0px 0px},clip]{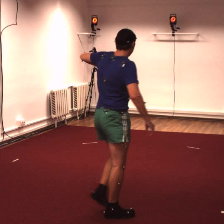}
    \includegraphics[height=0.15\linewidth,trim={0px 0px 0px 0px},clip]{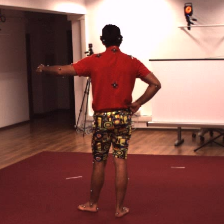}
    \includegraphics[height=0.15\linewidth,trim={0px 0px 0px 0px},clip]{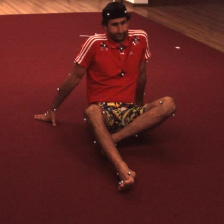}
    \includegraphics[height=0.15\linewidth,trim={0px 0px 0px 0px},clip]{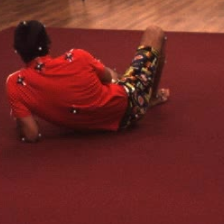}
    
    \includegraphics[height=0.15\linewidth,trim={100px 80px 100px 120px},clip]{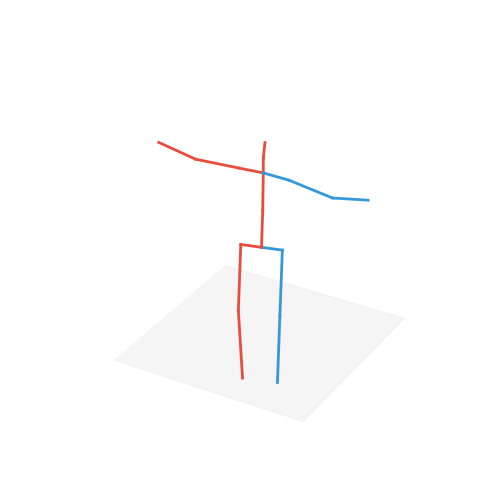}
    \includegraphics[height=0.15\linewidth,trim={100px 80px 100px 120px},clip]{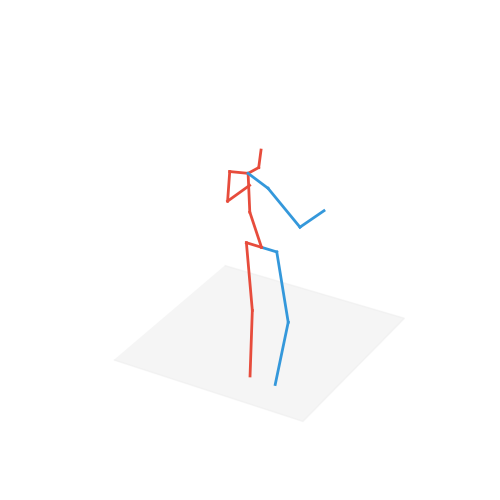}
    \includegraphics[height=0.15\linewidth,trim={100px 80px 100px 120px},clip]{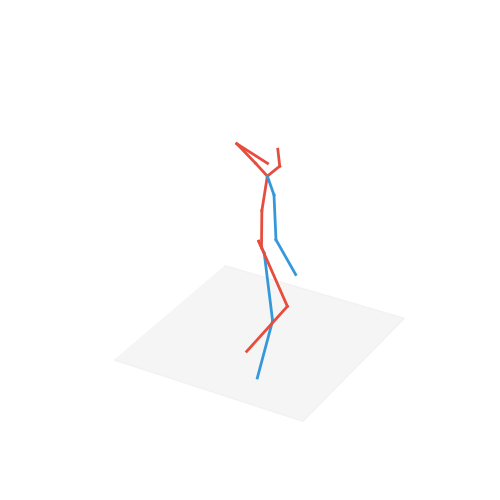}
    \includegraphics[height=0.15\linewidth,trim={100px 80px 100px 120px},clip]{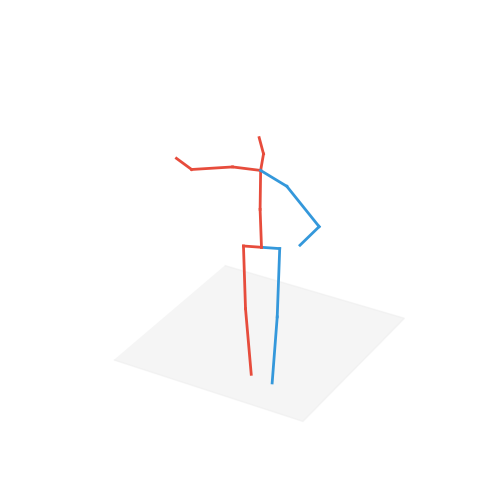}
    \includegraphics[height=0.15\linewidth,trim={100px 80px 100px 120px},clip]{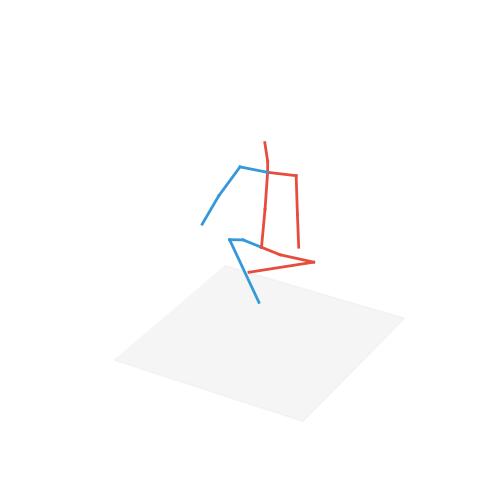}
    \includegraphics[height=0.15\linewidth,trim={100px 80px 100px 120px},clip]{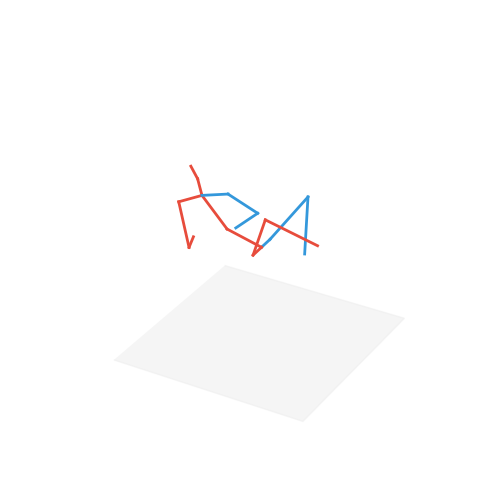}
    
    \includegraphics[height=0.15\linewidth,trim={100px 80px 100px 120px},clip]{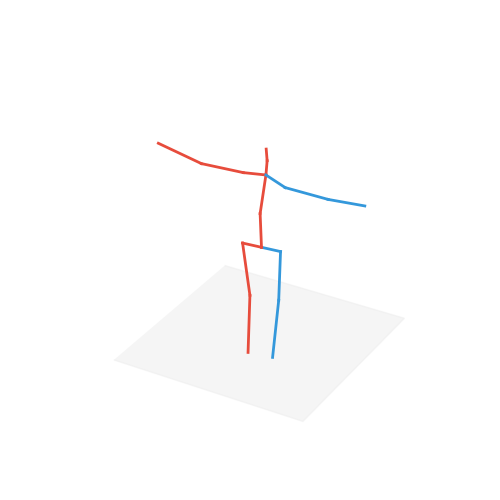}
    \includegraphics[height=0.15\linewidth,trim={100px 80px 100px 120px},clip]{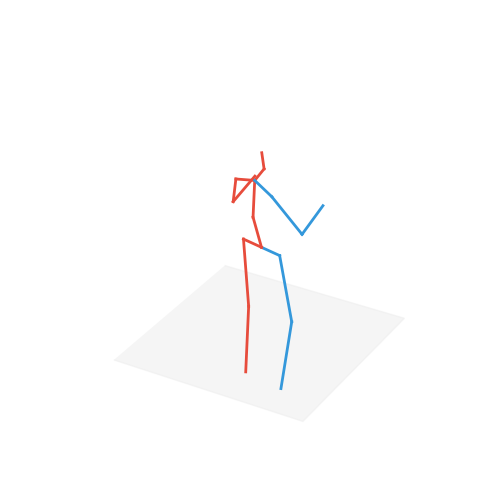}
    \includegraphics[height=0.15\linewidth,trim={100px 80px 100px 120px},clip]{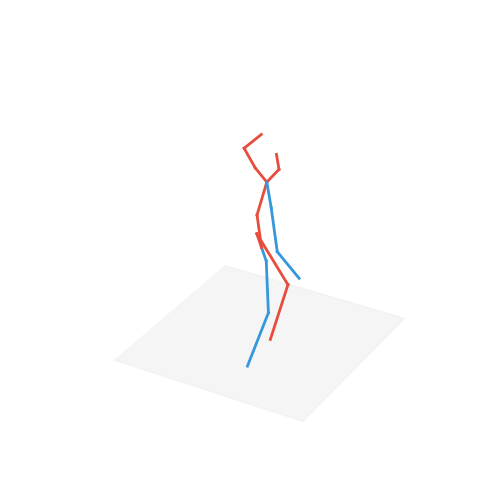}
    \includegraphics[height=0.15\linewidth,trim={100px 80px 100px 120px},clip]{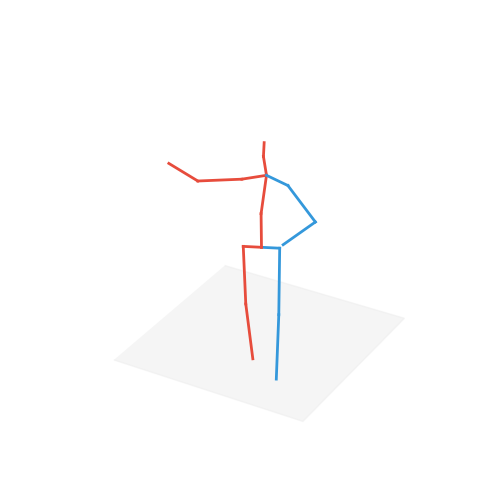}
    \includegraphics[height=0.15\linewidth,trim={100px 80px 100px 120px},clip]{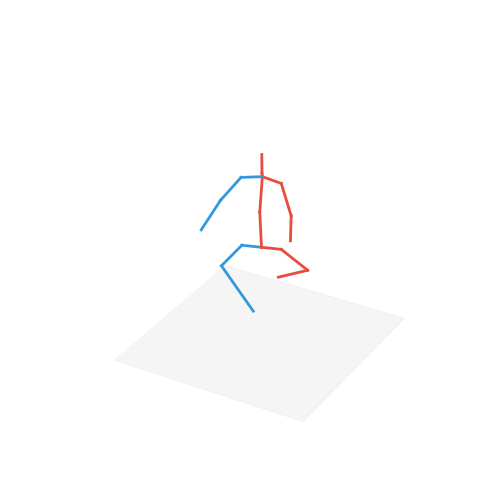}
    \includegraphics[height=0.15\linewidth,trim={100px 80px 100px 120px},clip]{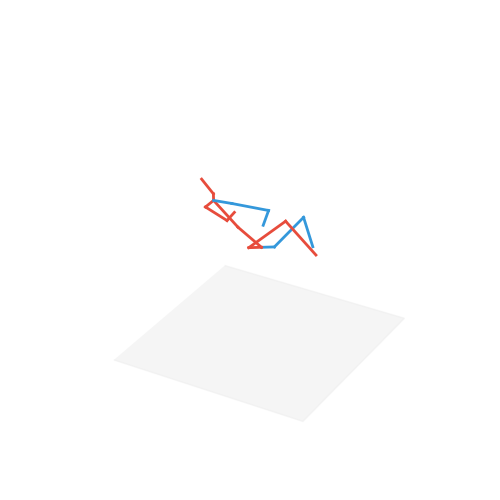}
    
    \includegraphics[height=0.15\linewidth,trim={100px 80px 100px 120px},clip]{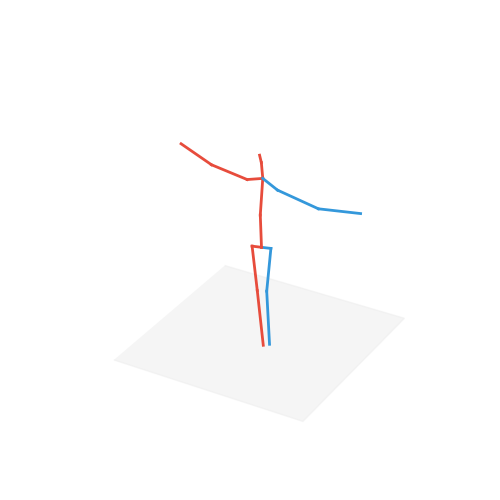}
    \includegraphics[height=0.15\linewidth,trim={100px 80px 100px 120px},clip]{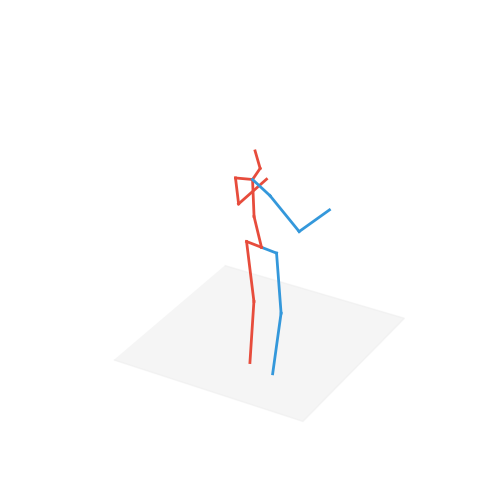}
    \includegraphics[height=0.15\linewidth,trim={100px 80px 100px 120px},clip]{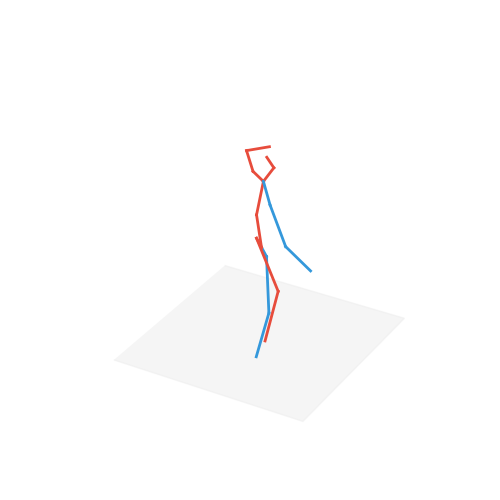}
    \includegraphics[height=0.15\linewidth,trim={100px 80px 100px 120px},clip]{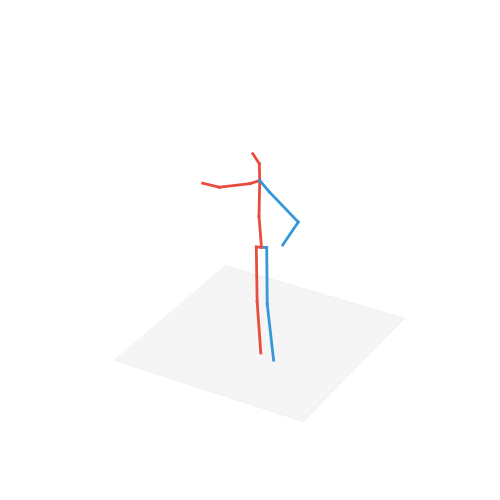}
    \includegraphics[height=0.15\linewidth,trim={100px 80px 100px 120px},clip]{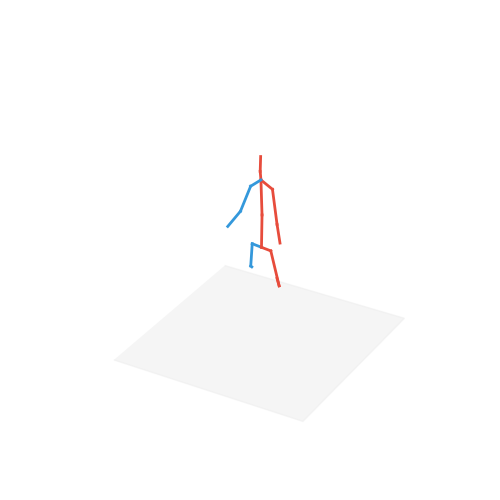}
    \includegraphics[height=0.15\linewidth,trim={100px 80px 100px 120px},clip]{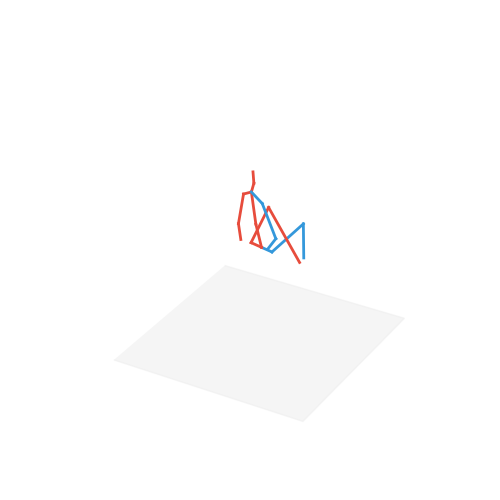}

    \caption{\textbf{Examples of predictions.} We show predictions on unseen test subjects (S9 and S11) of a model that was pre-trained using our self-supervised learning task and fine-tuned for 3D pose estimation. We show the test images on the first row, the ground truth poses on the second row, the predictions of a model fine-tuned on all training subjects on the third row, and the predictions of a model fine-tuned only on subject S1 on the last row.}
    \label{fig:pose_preds}
\end{figure}

The results and the comparisons to prior work are provided in Table~\ref{tab:comparison}. 
We observe that our fine-tuned network generalizes well and outperforms the relevant prior work \cite{rhodin2018learning,rhodin2018unsupervised,mitra2020multiview} by a considerable margin. Note that our ResNet-18 is the only method besides \cite{rhodin2018unsupervised} that does not rely on any prior supervision.
Interestingly, our model with substituted backgrounds performs better than our model with removed backgrounds in protocol \textbf{All} and worse in protocol \textbf{S1}. The difference is most severe in the scale sensitive metric MPJPE. We hypothesize that the backgrounds might be useful to learn the correct absolute scale, given a number of different training subjects. 
We also list methods that make use of large amounts of additional labelled 2D pose data. Using all the training subjects for transfer learning, we even outperform the weakly supervised method \cite{chen2019weakly}. We show some qualitative predictions for the two test subjects in Fig.~\ref{fig:pose_preds}.

\begin{figure}[t]
    \centering
    \includegraphics[width=.8\linewidth]{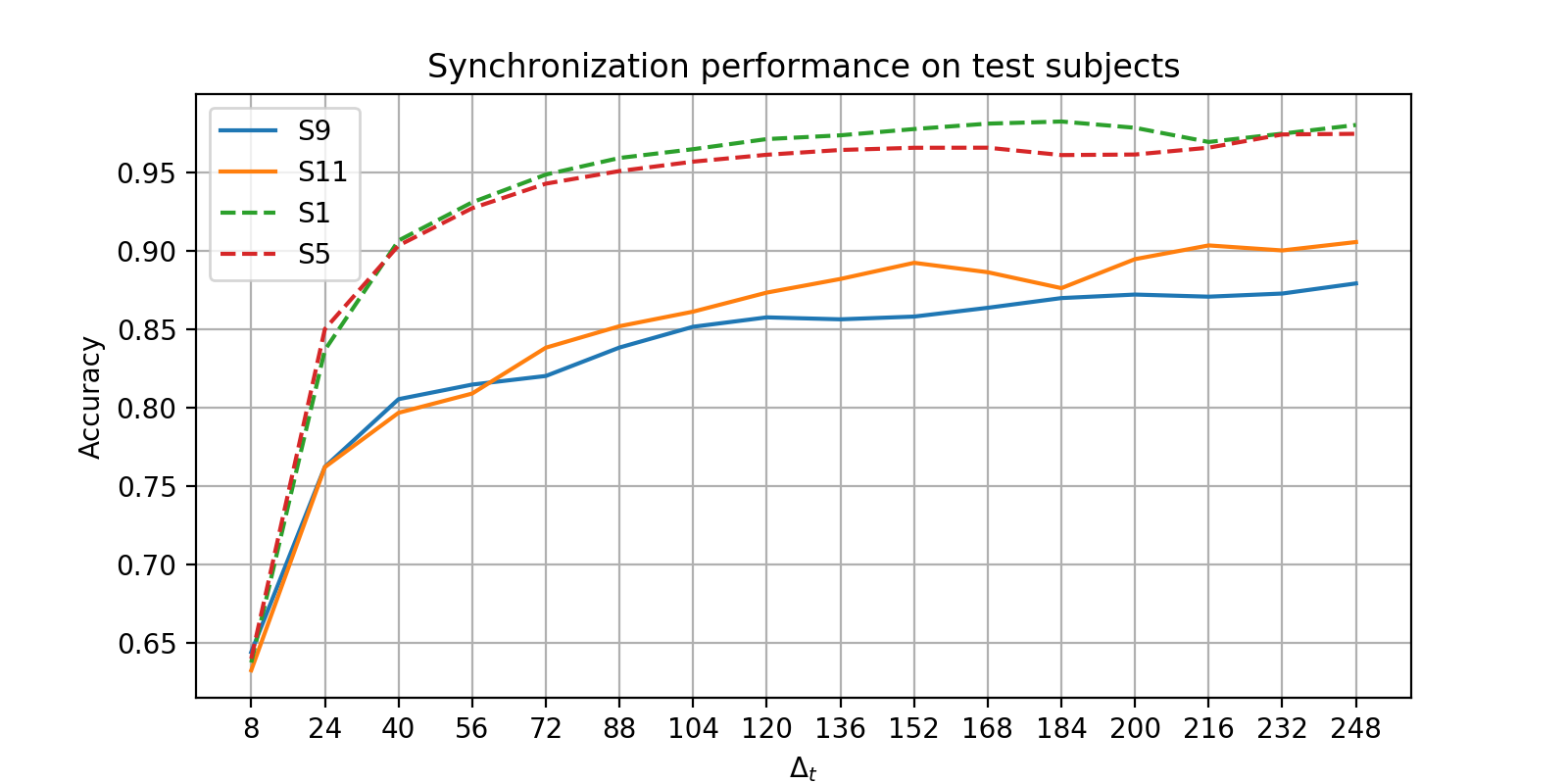}
    \caption{\textbf{Synchronization performance on unseen test subjects.} We evaluate the pre-trained synchronization network on test subjects S9 and S11. Unsynchronized pairs are created with a time difference sampled from the interval $[\Delta_t-7, \Delta_t+8]$. We also show the accuracy on two training subjects S1 and S5.}
    \label{fig:sync_perf}
\end{figure}

\subsection{Evaluation of the synchronization task}

\noindent \textbf{Generalization across subjects.} We qualitatively evaluate how well our model generalizes to new subjects on the synchronization prediction task. The idea is to look for synchronized frames, where the two frames contain different subjects. A query image $x^{(q)}$ of a previously unseen test subject is chosen. We look for the training image $x^{(a)}$ of each training subject that maximizes the predicted probability of synchronization, \ie, $a=\argmax_i F_1((x^{(q)}, x^{(i)}))$. Note that our network was not explicitly trained for this task and that it only received pairs of frames as input containing the same subject during training.
\begin{figure}[!t]
    \centering
    \includegraphics[height=0.16\linewidth,trim={0px 0px 0px 0px},clip]{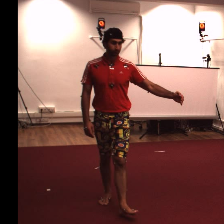}
    \includegraphics[height=0.16\linewidth,trim={0px 0px 0px 0px},clip]{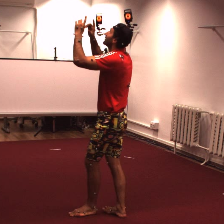}
    \includegraphics[height=0.16\linewidth,trim={0px 0px 0px 0px},clip]{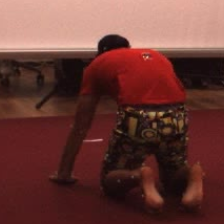}    
    \includegraphics[height=0.16\linewidth,trim={0px 0px 0px 0px},clip]{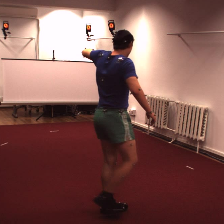}
    \includegraphics[height=0.16\linewidth,trim={0px 0px 0px 0px},clip]{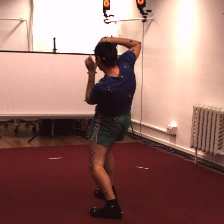}
    \includegraphics[height=0.16\linewidth,trim={0px 0px 0px 0px},clip]{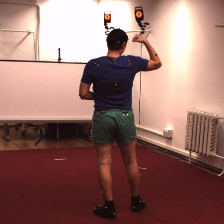}\\
    \includegraphics[height=0.16\linewidth,trim={0px 0px 0px 0px},clip]{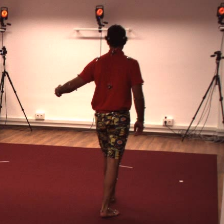}
    \includegraphics[height=0.16\linewidth,trim={0px 0px 0px 0px},clip]{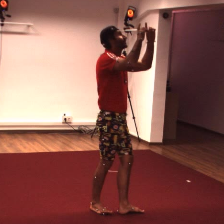}
    \includegraphics[height=0.16\linewidth,trim={0px 0px 0px 0px},clip]{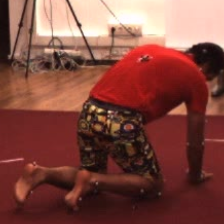}
    \includegraphics[height=0.16\linewidth,trim={0px 0px 0px 0px},clip]{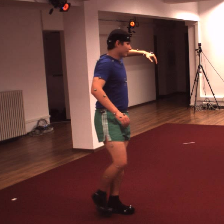}
    \includegraphics[height=0.16\linewidth,trim={0px 0px 0px 0px},clip]{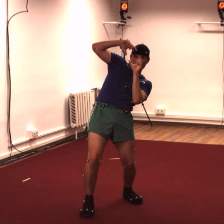}
    \includegraphics[height=0.16\linewidth,trim={0px 0px 0px 0px},clip]{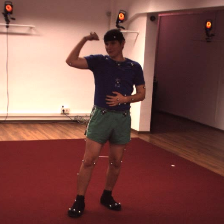}
    \caption{\textbf{Synchronization retrievals on test subjects.} Query images are shown on the top row and the corresponding predicted synchronized frames are shown in the row below.}
    \label{fig:sync_test}
\end{figure}

\begin{figure}[t]
    \centering
    \includegraphics[height=0.155\linewidth,trim={0px 0px 0px 0px},clip]{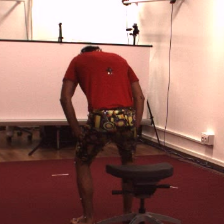}
    \includegraphics[height=0.155\linewidth,trim={0px 0px 0px 0px},clip]{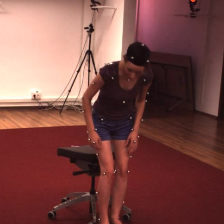}
    \includegraphics[height=0.155\linewidth,trim={0px 0px 0px 0px},clip]{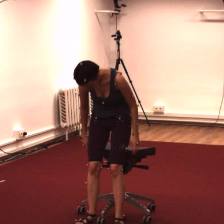}
    \includegraphics[height=0.155\linewidth,trim={0px 0px 0px 0px},clip]{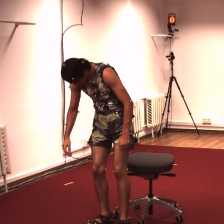}
    \includegraphics[height=0.155\linewidth,trim={0px 0px 0px 0px},clip]{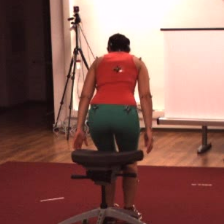}
    \includegraphics[height=0.155\linewidth,trim={0px 0px 0px 0px},clip]{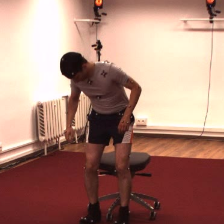}\\
    \includegraphics[height=0.155\linewidth,trim={0px 0px 0px 0px},clip]{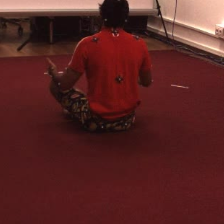}
    \includegraphics[height=0.155\linewidth,trim={0px 0px 0px 0px},clip]{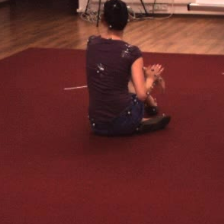}
    \includegraphics[height=0.155\linewidth,trim={0px 0px 0px 0px},clip]{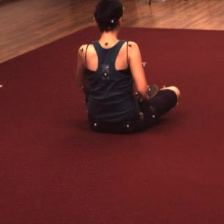}
    \includegraphics[height=0.155\linewidth,trim={0px 0px 0px 0px},clip]{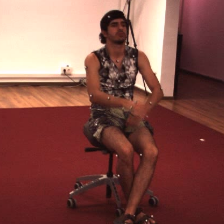}
    \includegraphics[height=0.155\linewidth,trim={0px 0px 0px 0px},clip]{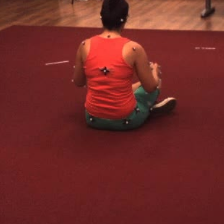}
    \includegraphics[height=0.155\linewidth,trim={0px 0px 0px 0px},clip]{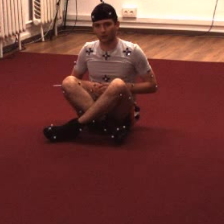}\\
    \includegraphics[height=0.155\linewidth,trim={0px 0px 0px 0px},clip]{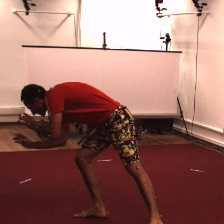}
    \includegraphics[height=0.155\linewidth,trim={0px 0px 0px 0px},clip]{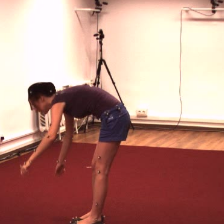}
    \includegraphics[height=0.155\linewidth,trim={0px 0px 0px 0px},clip]{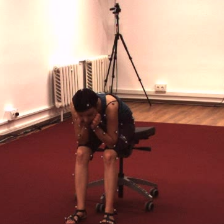}
    \includegraphics[height=0.155\linewidth,trim={0px 0px 0px 0px},clip]{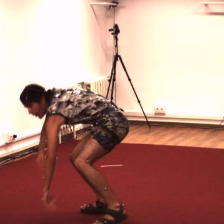}
    \includegraphics[height=0.155\linewidth,trim={0px 0px 0px 0px},clip]{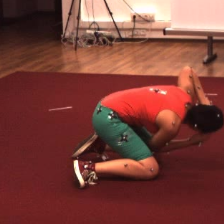}
    \includegraphics[height=0.155\linewidth,trim={0px 0px 0px 0px},clip]{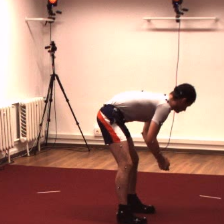}\\
    \includegraphics[height=0.155\linewidth,trim={0px 0px 0px 0px},clip]{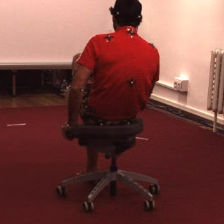}
    \includegraphics[height=0.155\linewidth,trim={0px 0px 0px 0px},clip]{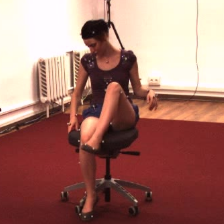}
    \includegraphics[height=0.155\linewidth,trim={0px 0px 0px 0px},clip]{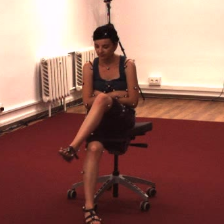}
    \includegraphics[height=0.155\linewidth,trim={0px 0px 0px 0px},clip]{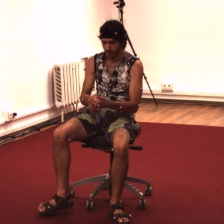}
    \includegraphics[height=0.155\linewidth,trim={0px 0px 0px 0px},clip]{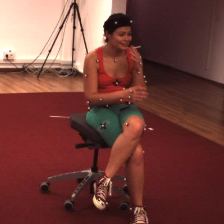}
    \includegraphics[height=0.155\linewidth,trim={0px 0px 0px 0px},clip]{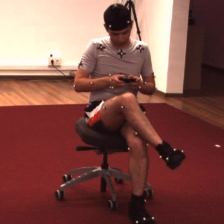}\\
    \includegraphics[height=0.155\linewidth,trim={0px 0px 0px 0px},clip]{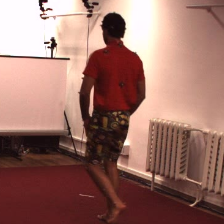}
    \includegraphics[height=0.155\linewidth,trim={0px 0px 0px 0px},clip]{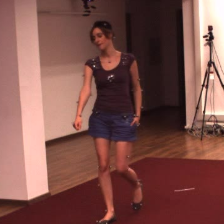}
    \includegraphics[height=0.155\linewidth,trim={0px 0px 0px 0px},clip]{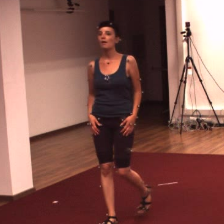}
    \includegraphics[height=0.155\linewidth,trim={0px 0px 0px 0px},clip]{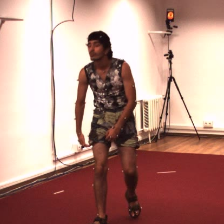}
    \includegraphics[height=0.155\linewidth,trim={0px 0px 0px 0px},clip]{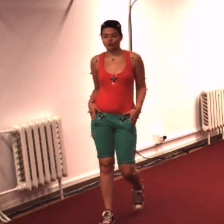}
    \includegraphics[height=0.155\linewidth,trim={0px 0px 0px 0px},clip]{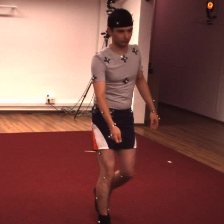}
    \caption{\textbf{Generalization across subjects.} We investigate if a network trained on our self-supervised synchronization task generalizes across different subjects. The left-most column shows the query image from test subject S9. The remaining columns shows images of training subjects S1, S5, S6, S7, and S8, respectively, for which our model predicts the highest probability of being synchronized.}
    \label{fig:sync_gen}
\end{figure}

We show example retrievals in Fig.~\ref{fig:sync_gen}. We can observe that the retrieved images often show very similar poses to the query image. The retrieved images also cover different viewpoints. The method generalizes surprisingly well across subjects and manages to associate similar poses across subjects. This indicates a certain degree of invariance to the person appearance. As discussed in the Introduction, such an invariance is welcome in 3D human pose estimation and could explain the good performance in the transfer experiments.\\
\noindent \textbf{Synchronization performance on test subjects.} 
We evaluate the performance of the synchronization network also on the test subjects. To this end, we sample synchronized and unsynchronized pairs in equal proportion with varying time gaps. We plot the accuracy in Fig.~\ref{fig:sync_perf}. Note that we use all the frames in the test examples in this case. As expected we see an increase in performance with a larger temporal gap between the two frames. We also show some qualitative samples of synchronization retrievals in Fig.~\ref{fig:sync_test}.

\section{Conclusions}

We propose a novel self-supervised learning method to tackle monocular 3D human pose estimation. Our method delivers a high performance without requiring large manually labeled data sets (\eg, with 2D or 3D poses).
To avoid such detailed annotation, we exploit a novel self-supervised task that leads to a representation that supports 3D pose estimation. Our task is to detect when two views have been captured at the same time (and thus the scene is related by a rigid transformation) and when they are horizontally flipped with respect to one another. We show on the well-known Human3.6M data set that these two objectives build features that generalize across subjects and are highly sensitive to small pose variations.

\noindent \textbf{Acknowledgements.} This work was supported by grant 169622 of the Swiss National Science Foundation (SNSF).

\FloatBarrier

\bibliographystyle{splncs}
\bibliography{egbib}

\begin{thebibliography}{10}

\bibitem{rosenhahn2008human}
Rosenhahn, B., Klette, R., Metaxas, D.:
\newblock Human motion: Understanding, Modeling, Capture.
\newblock Springer (2008)

\bibitem{h36m_pami}
Ionescu, C., Papava, D., Olaru, V., Sminchisescu, C.:
\newblock Human3.6m: Large scale datasets and predictive methods for 3d human
  sensing in natural environments.
\newblock IEEE Transactions on Pattern Analysis and Machine Intelligence
  \textbf{36} (2014)  1325--1339

\bibitem{doersch2015unsupervised}
Doersch, C., Gupta, A., Efros, A.A.:
\newblock Unsupervised visual representation learning by context prediction.
\newblock In: Proceedings of the IEEE International Conference on Computer
  Vision. (2015)  1422--1430

\bibitem{nityananda2017stereopsis}
Nityananda, V., Read, J.C.:
\newblock Stereopsis in animals: evolution, function and mechanisms.
\newblock Journal of Experimental Biology \textbf{220} (2017)  2502--2512

\bibitem{harland2012jumping}
Harland, D.P., Li, D., Jackson, R.R.:
\newblock How jumping spiders see the world.
\newblock Oxford University Press (2012)

\bibitem{pathak2016context}
Pathak, D., Krahenbuhl, P., Donahue, J., Darrell, T., Efros, A.A.:
\newblock Context encoders: Feature learning by inpainting.
\newblock In: Proceedings of the IEEE conference on computer vision and pattern
  recognition. (2016)  2536--2544

\bibitem{zhang2016colorful}
Zhang, R., Isola, P., Efros, A.A.:
\newblock Colorful image colorization.
\newblock In: European conference on computer vision, Springer (2016)  649--666

\bibitem{noroozi2016unsupervised}
Noroozi, M., Favaro, P.:
\newblock Unsupervised learning of visual representations by solving jigsaw
  puzzles.
\newblock In: European Conference on Computer Vision, Springer (2016)  69--84

\bibitem{chen2020simple}
Chen, T., Kornblith, S., Norouzi, M., Hinton, G.:
\newblock A simple framework for contrastive learning of visual
  representations.
\newblock arXiv preprint arXiv:2002.05709 (2020)

\bibitem{he2020momentum}
He, K., Fan, H., Wu, Y., Xie, S., Girshick, R.:
\newblock Momentum contrast for unsupervised visual representation learning.
\newblock In: Proceedings of the IEEE/CVF Conference on Computer Vision and
  Pattern Recognition. (2020)  9729--9738

\bibitem{caron2018deep}
Caron, M., Bojanowski, P., Joulin, A., Douze, M.:
\newblock Deep clustering for unsupervised learning of visual features.
\newblock In: Proceedings of the European Conference on Computer Vision (ECCV).
  (2018)  132--149

\bibitem{jenni2020steering}
Jenni, S., Jin, H., Favaro, P.:
\newblock Steering self-supervised feature learning beyond local pixel
  statistics.
\newblock In: Proceedings of the IEEE/CVF Conference on Computer Vision and
  Pattern Recognition. (2020)  6408--6417

\bibitem{gidaris2018unsupervised}
Gidaris, S., Singh, P., Komodakis, N.:
\newblock Unsupervised representation learning by predicting image rotations.
\newblock arXiv preprint arXiv:1803.07728 (2018)

\bibitem{zhang2019aet}
Zhang, L., Qi, G.J., Wang, L., Luo, J.:
\newblock Aet vs. aed: Unsupervised representation learning by auto-encoding
  transformations rather than data.
\newblock In: Proceedings of the IEEE Conference on Computer Vision and Pattern
  Recognition. (2019)  2547--2555

\bibitem{wu2020unsupervised}
Wu, S., Rupprecht, C., Vedaldi, A.:
\newblock Unsupervised learning of probably symmetric deformable 3d objects
  from images in the wild.
\newblock In: Proceedings of the IEEE/CVF Conference on Computer Vision and
  Pattern Recognition. (2020)  1--10

\bibitem{nguyen2019hologan}
Nguyen-Phuoc, T., Li, C., Theis, L., Richardt, C., Yang, Y.L.:
\newblock Hologan: Unsupervised learning of 3d representations from natural
  images.
\newblock In: Proceedings of the IEEE International Conference on Computer
  Vision. (2019)  7588--7597

\bibitem{szabo2019unsupervised}
Szab{\'o}, A., Meishvili, G., Favaro, P.:
\newblock Unsupervised generative 3d shape learning from natural images.
\newblock arXiv preprint arXiv:1910.00287 (2019)

\bibitem{agarwala2005panoramic}
Agarwala, A., Zheng, K.C., Pal, C., Agrawala, M., Cohen, M., Curless, B.,
  Salesin, D., Szeliski, R.:
\newblock Panoramic video textures.
\newblock In: ACM SIGGRAPH 2005 Papers.
\newblock (2005)  821--827

\bibitem{sand2004video}
Sand, P., Teller, S.:
\newblock Video matching.
\newblock ACM Transactions on Graphics (TOG) \textbf{23} (2004)  592--599

\bibitem{tuytelaars2004synchronizing}
Tuytelaars, T., Van~Gool, L.:
\newblock Synchronizing video sequences.
\newblock In: Proceedings of the 2004 IEEE Computer Society Conference on
  Computer Vision and Pattern Recognition, 2004. CVPR 2004. Volume~1., IEEE
  (2004)  I--I

\bibitem{wieschollek2017learning}
Wieschollek, P., Freeman, I., Lensch, H.P.:
\newblock Learning robust video synchronization without annotations.
\newblock In: 2017 16th IEEE International Conference on Machine Learning and
  Applications (ICMLA), IEEE (2017)  92--100

\bibitem{liang2017synchronization}
Liang, J., Huang, P., Chen, J., Hauptmann, A.:
\newblock Synchronization for multi-perspective videos in the wild.
\newblock In: 2017 IEEE International Conference on Acoustics, Speech and
  Signal Processing (ICASSP), IEEE (2017)  1592--1596

\bibitem{andriluka20142d}
Andriluka, M., Pishchulin, L., Gehler, P., Schiele, B.:
\newblock 2d human pose estimation: New benchmark and state of the art
  analysis.
\newblock In: Proceedings of the IEEE Conference on computer Vision and Pattern
  Recognition. (2014)  3686--3693

\bibitem{li20143d}
Li, S., Chan, A.B.:
\newblock 3d human pose estimation from monocular images with deep
  convolutional neural network.
\newblock In: Asian Conference on Computer Vision, Springer (2014)  332--347

\bibitem{li2015maximum}
Li, S., Zhang, W., Chan, A.B.:
\newblock Maximum-margin structured learning with deep networks for 3d human
  pose estimation.
\newblock In: Proceedings of the IEEE international conference on computer
  vision. (2015)  2848--2856

\bibitem{tekin2016structured}
Tekin, B., Katircioglu, I., Salzmann, M., Lepetit, V., Fua, P.:
\newblock Structured prediction of 3d human pose with deep neural networks.
\newblock arXiv preprint arXiv:1605.05180 (2016)

\bibitem{zhou2016deep}
Zhou, X., Sun, X., Zhang, W., Liang, S., Wei, Y.:
\newblock Deep kinematic pose regression.
\newblock In: European Conference on Computer Vision, Springer (2016)  186--201

\bibitem{tekin2017learning}
Tekin, B., M{\'a}rquez-Neila, P., Salzmann, M., Fua, P.:
\newblock Learning to fuse 2d and 3d image cues for monocular body pose
  estimation.
\newblock In: Proceedings of the IEEE International Conference on Computer
  Vision. (2017)  3941--3950

\bibitem{pavlakos2017harvesting}
Pavlakos, G., Zhou, X., Derpanis, K.G., Daniilidis, K.:
\newblock Harvesting multiple views for marker-less 3d human pose annotations.
\newblock In: Proceedings of the IEEE conference on computer vision and pattern
  recognition. (2017)  6988--6997

\bibitem{pavlakos2017coarse}
Pavlakos, G., Zhou, X., Derpanis, K.G., Daniilidis, K.:
\newblock Coarse-to-fine volumetric prediction for single-image 3d human pose.
\newblock In: Proceedings of the IEEE Conference on Computer Vision and Pattern
  Recognition. (2017)  7025--7034

\bibitem{mehta2017monocular}
Mehta, D., Rhodin, H., Casas, D., Fua, P., Sotnychenko, O., Xu, W., Theobalt,
  C.:
\newblock Monocular 3d human pose estimation in the wild using improved cnn
  supervision.
\newblock In: 2017 international conference on 3D vision (3DV), IEEE (2017)
  506--516

\bibitem{sun2018integral}
Sun, X., Xiao, B., Wei, F., Liang, S., Wei, Y.:
\newblock Integral human pose regression.
\newblock In: Proceedings of the European Conference on Computer Vision (ECCV).
  (2018)  529--545

\bibitem{popa2017deep}
Popa, A.I., Zanfir, M., Sminchisescu, C.:
\newblock Deep multitask architecture for integrated 2d and 3d human sensing.
\newblock In: Proceedings of the IEEE Conference on Computer Vision and Pattern
  Recognition. (2017)  6289--6298

\bibitem{tome2017lifting}
Tome, D., Russell, C., Agapito, L.:
\newblock Lifting from the deep: Convolutional 3d pose estimation from a single
  image.
\newblock In: Proceedings of the IEEE Conference on Computer Vision and Pattern
  Recognition. (2017)  2500--2509

\bibitem{mehta2017vnect}
Mehta, D., Sridhar, S., Sotnychenko, O., Rhodin, H., Shafiei, M., Seidel, H.P.,
  Xu, W., Casas, D., Theobalt, C.:
\newblock Vnect: Real-time 3d human pose estimation with a single rgb camera.
\newblock ACM Transactions on Graphics (TOG) \textbf{36} (2017)  1--14

\bibitem{rogez2017lcr}
Rogez, G., Weinzaepfel, P., Schmid, C.:
\newblock Lcr-net: Localization-classification-regression for human pose.
\newblock In: Proceedings of the IEEE Conference on Computer Vision and Pattern
  Recognition. (2017)  3433--3441

\bibitem{dabral2018learning}
Dabral, R., Mundhada, A., Kusupati, U., Afaque, S., Sharma, A., Jain, A.:
\newblock Learning 3d human pose from structure and motion.
\newblock In: Proceedings of the European Conference on Computer Vision (ECCV).
  (2018)  668--683

\bibitem{martinez2017simple}
Martinez, J., Hossain, R., Romero, J., Little, J.J.:
\newblock A simple yet effective baseline for 3d human pose estimation.
\newblock In: Proceedings of the IEEE International Conference on Computer
  Vision. (2017)  2640--2649

\bibitem{zhou2017towards}
Zhou, X., Huang, Q., Sun, X., Xue, X., Wei, Y.:
\newblock Towards 3d human pose estimation in the wild: a weakly-supervised
  approach.
\newblock In: Proceedings of the IEEE International Conference on Computer
  Vision. (2017)  398--407

\bibitem{moreno20173d}
Moreno-Noguer, F.:
\newblock 3d human pose estimation from a single image via distance matrix
  regression.
\newblock In: Proceedings of the IEEE Conference on Computer Vision and Pattern
  Recognition. (2017)  2823--2832

\bibitem{rayat2018exploiting}
Rayat Imtiaz~Hossain, M., Little, J.J.:
\newblock Exploiting temporal information for 3d human pose estimation.
\newblock In: Proceedings of the European Conference on Computer Vision (ECCV).
  (2018)  68--84

\bibitem{fang2018learning}
Fang, H.S., Xu, Y., Wang, W., Liu, X., Zhu, S.C.:
\newblock Learning pose grammar to encode human body configuration for 3d pose
  estimation.
\newblock In: Thirty-Second AAAI Conference on Artificial Intelligence. (2018)

\bibitem{chen20173d}
Chen, C.H., Ramanan, D.:
\newblock 3d human pose estimation= 2d pose estimation+ matching.
\newblock In: Proceedings of the IEEE Conference on Computer Vision and Pattern
  Recognition. (2017)  7035--7043

\bibitem{zhao2019semantic}
Zhao, L., Peng, X., Tian, Y., Kapadia, M., Metaxas, D.N.:
\newblock Semantic graph convolutional networks for 3d human pose regression.
\newblock In: Proceedings of the IEEE Conference on Computer Vision and Pattern
  Recognition. (2019)  3425--3435

\bibitem{sharma2019monocular}
Sharma, S., Varigonda, P.T., Bindal, P., Sharma, A., Jain, A.:
\newblock Monocular 3d human pose estimation by generation and ordinal ranking.
\newblock In: Proceedings of the IEEE International Conference on Computer
  Vision. (2019)  2325--2334

\bibitem{wang20193d}
Wang, K., Lin, L., Jiang, C., Qian, C., Wei, P.:
\newblock 3d human pose machines with self-supervised learning.
\newblock IEEE Transactions on Pattern Analysis and Machine Intelligence
  \textbf{42} (2019)  1069--1082

\bibitem{chen2016synthesizing}
Chen, W., Wang, H., Li, Y., Su, H., Wang, Z., Tu, C., Lischinski, D., Cohen-Or,
  D., Chen, B.:
\newblock Synthesizing training images for boosting human 3d pose estimation.
\newblock In: 2016 Fourth International Conference on 3D Vision (3DV), IEEE
  (2016)  479--488

\bibitem{rogez2016mocap}
Rogez, G., Schmid, C.:
\newblock Mocap-guided data augmentation for 3d pose estimation in the wild.
\newblock In: Advances in neural information processing systems. (2016)
  3108--3116

\bibitem{varol2017learning}
Varol, G., Romero, J., Martin, X., Mahmood, N., Black, M.J., Laptev, I.,
  Schmid, C.:
\newblock Learning from synthetic humans.
\newblock In: Proceedings of the IEEE Conference on Computer Vision and Pattern
  Recognition. (2017)  109--117

\bibitem{kocabas2019self}
Kocabas, M., Karagoz, S., Akbas, E.:
\newblock Self-supervised learning of 3d human pose using multi-view geometry.
\newblock In: Proceedings of the IEEE Conference on Computer Vision and Pattern
  Recognition. (2019)  1077--1086

\bibitem{rhodin2018learning}
Rhodin, H., Sp{\"o}rri, J., Katircioglu, I., Constantin, V., Meyer, F.,
  M{\"u}ller, E., Salzmann, M., Fua, P.:
\newblock Learning monocular 3d human pose estimation from multi-view images.
\newblock In: Proceedings of the IEEE Conference on Computer Vision and Pattern
  Recognition. (2018)  8437--8446

\bibitem{chen2019weakly}
Chen, X., Lin, K.Y., Liu, W., Qian, C., Lin, L.:
\newblock Weakly-supervised discovery of geometry-aware representation for 3d
  human pose estimation.
\newblock In: Proceedings of the IEEE Conference on Computer Vision and Pattern
  Recognition. (2019)  10895--10904

\bibitem{chen2019unsupervised}
Chen, C.H., Tyagi, A., Agrawal, A., Drover, D., Stojanov, S., Rehg, J.M.:
\newblock Unsupervised 3d pose estimation with geometric self-supervision.
\newblock In: Proceedings of the IEEE Conference on Computer Vision and Pattern
  Recognition. (2019)  5714--5724

\bibitem{kanazawa2018end}
Kanazawa, A., Black, M.J., Jacobs, D.W., Malik, J.:
\newblock End-to-end recovery of human shape and pose.
\newblock In: Proceedings of the IEEE Conference on Computer Vision and Pattern
  Recognition. (2018)  7122--7131

\bibitem{wandt2019repnet}
Wandt, B., Rosenhahn, B.:
\newblock Repnet: Weakly supervised training of an adversarial reprojection
  network for 3d human pose estimation.
\newblock In: Proceedings of the IEEE conference on computer vision and pattern
  recognition. (2019)  7782--7791

\bibitem{pavllo20193d}
Pavllo, D., Feichtenhofer, C., Grangier, D., Auli, M.:
\newblock 3d human pose estimation in video with temporal convolutions and
  semi-supervised training.
\newblock In: Proceedings of the IEEE Conference on Computer Vision and Pattern
  Recognition. (2019)  7753--7762

\bibitem{rhodin2018unsupervised}
Rhodin, H., Salzmann, M., Fua, P.:
\newblock Unsupervised geometry-aware representation for 3d human pose
  estimation.
\newblock In: Proceedings of the European Conference on Computer Vision (ECCV).
  (2018)  750--767

\bibitem{mitra2020multiview}
Mitra, R., Gundavarapu, N.B., Sharma, A., Jain, A.:
\newblock Multiview-consistent semi-supervised learning for 3d human pose
  estimation.
\newblock In: Proceedings of the IEEE/CVF Conference on Computer Vision and
  Pattern Recognition. (2020)  6907--6916

\bibitem{he2016deep}
He, K., Zhang, X., Ren, S., Sun, J.:
\newblock Deep residual learning for image recognition.
\newblock In: Proceedings of the IEEE conference on computer vision and pattern
  recognition. (2016)  770--778

\bibitem{loshchilov2018decoupled}
Loshchilov, I., Hutter, F.:
\newblock Fixing weight decay regularization in adam.
\newblock arXiv preprint arXiv:1711.05101 (2017)

\bibitem{loshchilov2016sgdr}
Loshchilov, I., Hutter, F.:
\newblock Sgdr: Stochastic gradient descent with warm restarts.
\newblock arXiv preprint arXiv:1608.03983 (2016)

\end{thebibliography}

\end{document}